\documentclass[letterpaper, 10 pt, conference]{ieeeconf}  

\IEEEoverridecommandlockouts                              
\usepackage[utf8]{inputenc}
\usepackage[T1]{fontenc}
\usepackage{graphicx}
\usepackage{algorithm}
\usepackage{algpseudocode}
\usepackage{float}
\usepackage{amsmath,amsfonts}
\usepackage{array}
\usepackage{booktabs}
\usepackage{subcaption}
\usepackage{url}
\usepackage{xcolor}

\title{\LARGE \bf
Poisoning Attacks on Federated Learning for Autonomous Driving
}

\author{Sonakshi Garg$^{1,2}$, Hugo Jönsson$^{2,3}$, Gustav Kalander$^{2,4}$, Axel Nilsson$^{2,3}$, \\ Bhhaanu Pirange$^{2,5}$, Viktor Valadi$^{2,6}$, and Johan Östman$^{2}$
\thanks{$^{1}$ Ume\r{a} University}%
\thanks{$^{2}$ AI Sweden}%
\thanks{$^{3}$ Royal Institute of Technology}%
\thanks{$^{4}$ Chalmers University of Technology}%
\thanks{$^{5}$ Dakota State University}%
\thanks{$^{6}$ Scaleout Systems}%
}

\begin{document}

\maketitle
\thispagestyle{empty}
\pagestyle{empty}

\begin{abstract}

Federated Learning (FL) is a decentralized learning paradigm, enabling parties to collaboratively train models while keeping their data confidential. 
Within autonomous driving, it brings the potential of reducing data storage costs, reducing bandwidth requirements, and to accelerate the learning.
FL is, however, susceptible to poisoning attacks.
In this paper, we introduce two novel poisoning attacks on FL tailored to regression tasks within autonomous driving: FLStealth and Off-Track Attack (OTA).
FLStealth, an untargeted attack, aims at providing model updates that deteriorate the global model performance while appearing benign.
OTA, on the other hand, is a targeted attack with the objective to change the global model's behavior when exposed to a certain trigger.
We demonstrate the effectiveness of our attacks by conducting comprehensive experiments pertaining to the task of vehicle trajectory prediction.
In particular, we show that, among five different untargeted attacks, FLStealth is the most successful at bypassing the considered defenses employed by the server.
For OTA, we demonstrate the inability of common defense strategies to mitigate the attack, highlighting the critical need for new defensive mechanisms against targeted attacks within FL for autonomous driving.

\end{abstract}

\section{Introduction}

Machine learning models deployed in-car are typically trained centrally on vast amounts of collected data~\cite{bojarski2016end}. 
However, centrally stored data is subject to large costs and may be subject to privacy concerns in relation to, e.g., the GDPR~\cite{GDPR}.
Further, in the case of wireless data collection, the data transmission requires significant bandwidth.
To remedy these shortcomings, federated learning (FL) has been proposed as a potential solution.
The main idea of FL is to train machine learning models locally, thereby maintaining data confidentiality, and then aggregate the locally trained models centrally into a global model~\cite{mcmahan2017communication}. 
Several FL frameworks, tailored for autonomous driving, have recently been introduced~\cite{aparna2021steering,savazzi2021opportunities, nguyen2022deep}. 

Within the automotive sector, companies like Toyota and Ford are exploring FL solutions across various applications, e.g., object detection~\cite{FedLearnIR} and turn-signal prediction~\cite{TurnSigFed}. 
As vehicular networks are intrinsically dynamic, a recent direction of research also pertains to developing novel protocols for the selection of vehicle within the federation~\cite{OrchFed}. 
However, as control is moved from a central entity to the vehicles, new attack surfaces emerge.
For example, a given vehicle may manipulate their local model towards a malicious objective, referred to as a poisoning attack, which could ultimately result in traffic accidents.
Hence, in any FL application, it is imperative to provide defences against vehicles with devious intentions.
A common mitigation strategy to such attacks is to employ robust aggregation of local models where the impact of outliers is limited~\cite{blanchard2017machine, cao2020fltrust}.

From the adversary perspective, poisoning attacks on FL are commonly tailored towards classification problems~\cite{huang2020dynamic,sun2021data} with only a small number targeting regression problems~\cite{li2021backdoor,wang2023bandit}. 
However, regression tasks are common in autonomous driving, e.g., vehicle speed prediction, distance estimation, time-to-collision prediction, and vehicle trajectory prediction. 
Therefore, in this paper, we investigate poisoning attacks on FL for regression tasks within autonomous driving. 
We introduce two attacks coined \textsc{FLStealth} and Off-Track Attack (\textsc{OTA}).
The former is a general untargeted attack with the objective to deteriorate the global model performance whereas the latter is a backdoor attack tailored specifically to the problem of vehicle trajectory prediction.
We conduct an experimental study, using the Zenseact Open Dataset (ZOD)~\cite{alibeigi2023zenseact}, on the impact of untargeted attacks on vehicle trajectory prediction and to what extent common defenses are effective. 
Furthermore, by using \textsc{OTA}, we demonstrate that FL systems are vulnerable to targeted attacks and that they may significantly impact the behavior of the global model. 
Notably, common defense mechanism are largely inefficient against \textsc{OTA}.

\section{Preliminaries} \label{preli}

\subsection{Federated Learning}
Federated learning (FL) is a learning paradigm where multiple clients collaboratively train a model without revealing their local data~\cite{mcmahan2017communication}.
In particular, FL attempts to find a model $\theta^\star$ according to
\begin{equation}
\label{eq:fl}
    \theta^\star = \arg\min_{\theta} \frac{1}{n} \sum_{i=1}^n \mathbb{E}_{(x,y)\sim\mathcal{P}_i}\left[\ell(x,y;\theta) \right]
\end{equation}
where $n$ is the number of clients in the federation, $\ell(x,y;\theta)$ denotes the loss function, parameterized by the model $\theta$, evaluated on a sample $(x,y)$,  $\mathcal{P}_i$ denotes the local data distribution of client $i\in[n]$, and $\mathbb{E}[\cdot]$ is used for expectation.
Practically, the expectation is approximated locally by the sample average over a training dataset $D_i$ sampled from $\mathcal{P}_i$.

To solve~\eqref{eq:fl}, a server coordinates multiple clients over several rounds, each initiated by broadcasting a global model.
The server then collects a locally updated version of the broadcasted model from the clients and aggregates it into an updated global model.
This iterative procedure proceeds until the global model converges or a predefined number of training rounds is reached.

\subsection{Poisoning Attacks in Federated Learning}
FL is vulnerable to clients with malicious intent that may manipulate their local updates before sending it to the server, so-called poisoning attacks.
Such attacks are multifaceted and may be untargeted~\cite{mahloujifar2019universal,guerraoui2018hidden}, i.e., aim to deteriorate the global model performance, or targeted, i.e., alter the behavior of the global model on specific data samples~\cite{huang2020dynamic, xie2019dba,bagdasaryan2020backdoor}.
Poisoning attacks may be divided into data poisoning~\cite{munoz2017towards,tolpegin2020data,shejwalkar2022back} and model poisoning~\cite{guerraoui2018hidden,baruch2019little,fang2020local,shejwalkar2021manipulating} where the former alters the underlying dataset and the latter directly manipulates the model weights. 
It should be noted that any data poisoning attack can be replicated using a model poisoning attack.

Some common untargeted attacks include label flipping, gradient ascent attacks, and model shuffling.
In a label flipping attack, the attacker intentionally alters the labels within its dataset to prevent the global model from learning patterns in the data~\cite{biggio2012poisoning,tolpegin2020data}. 
In gradient ascent attacks, the attacker updates the model in the direction that maximizes the loss.
The model shuffling attack aims at shuffling the model parameters without notably changing the loss~\cite{yang2023model}.

Backdoor attacks typically rely on triggers injected in the data, causing the model to misbehave when exposed to the trigger~\cite{shejwalkar2022back, bagdasaryan2020backdoor}.
An example pertaining to street-sign detection is given in~\cite{gu2019badnets}.
Therein, a street-sign detector typically performs well but may incorrectly identify stop signs with a particular sticker as speed limit signs.
Such a behavior can be achieved by the following optimization procedure
\begin{equation}
    \theta^\star = \arg\min_\theta \sum_{(x,y) \in {D_{\mathrm{H}}}} \ell (x, y;\theta) + \sum_{(x,y)  \in D_{\mathrm{B}}} \ell (\mu (x, y);\theta)
\end{equation}
where $ D_{\mathrm{H}}$ denotes an honest dataset and $ D_{\mathrm{B}}$ a byzantine dataset to be used for the backdoor attack. 
Samples in $ D_{\mathrm{B}}$ are manipulated using some perturbation mechanism $\mu$ aligned with the backdoor objective. 
Notably, a backdoor attack aligns with the global objective on the honest dataset.

\subsection{Poisoning Mitigation Strategies in Federated Learning} 

Any convincing defensive mechanism should be able to handle an arbitrary attack.
For this reason, the byzantine threat model, allowing an attacker to directly alter the model weights to submit arbitrary updates, is prevalent.
Within byzantine resilient FL, there are two categories: robust aggregation~\cite{blanchard2017machine, yin2018byzantine, valadi2023fedval} and anomaly detection~\cite{xhemrishi2023fedgt}.
The former category is based on outlier mitigation, i.e., it relies on benign clients submitting similar models, whereas the latter category attempts to directly identify misbehaving clients.
In this paper, we shall focus on the former class of strategies.

A non-exhaustive list of robust aggregation techniques include \textsc{Krum}~\cite{blanchard2017machine}, \textsc{FLTrust}~\cite{cao2020fltrust}, \textsc{TrimmedMean}~\cite{yin2018byzantine}, PCA Defence~\cite{tolpegin2020data}, loss-function based rejection (\textsc{LFR})~\cite{fang2020local} and Loss Defence.
The first four methods relies on benign clients being similar to each other or to a server-based model whereas the last two removes clients that have a large impact on the global loss obtained via a server-based validation dataset.

\section{Novel Attacks on Regression Tasks} \label{method}
In this section, our threat model is defined and two novel attacks, pertaining to regression tasks in autonomous driving, are introduced.

\subsection{Threat Model}
We consider a federation with an honest-but-curious server and $n$ clients out of which $m<n$ are compromised.\footnote{We will refer to vehicles and clients interchangeably in the remainder of the paper.}
The $m$ malicious clients may collude to perform coordinated attacks.
Furthermore, the malicious clients may perform either data or model-poisoning attacks.

\subsection{\textsc{FLStealth}}
Based on the threat model, we now introduce a novel untargeted attack on federated regression tasks.
To circumvent any defensive efforts, the attack attempts to deteriorate the global model as much as possible while remaining stealthy.
This is achieved by creating two models, an honest and a byzantine, both initialized from the global model.
The attack is divided in two steps where the first accounts to training the honest model according as if the client was benign. 
Thereafter, the byzantine model is trained to maximize the loss while remaining close to the honest model. 
The resulting loss function of the byzantine model is given as
\begin{equation}
\label{eq:eq2}
    \ell_{\mathrm{FLStealth}}(x,y,\theta_{\mathrm{H}}; \theta_{\mathrm{B}}) = -\kappa \ell(x,y,\theta_\mathrm{B}) + \mathrm{MSE}(\theta_{\mathrm{H}}, \theta_{\mathrm{B}})
\end{equation}
where $\kappa \geq 0$ is a weighting constant, $\theta_i$, $i\in\lbrace \mathrm{H}, \mathrm{B} \rbrace$, denotes the honest and byzantine models, and $\mathrm{MSE}$ is the mean-squared error.
As can be seen, a lower $\kappa$ results in a byzantine model closer to the honest model.



\subsection{Off-Track Attack}
Next, we propose a novel backdoor attack crafted for vehicle trajectory prediction. 
It is based on the principle of triggers, as discussed in \cite{shejwalkar2022back, bagdasaryan2020backdoor}, but adapted towards the specific use-case of vehicle trajectory prediction.
For classification tasks, a backdoor attack can be as simple as flipping a class label.
However, for regression tasks there are no classes, hence, the target has to be altered differently.
In trajectory prediction, the target trajectory may be altered by slightly changing points resulting in an alternative path. 
The details are presented in Section~\ref{result}.

\section{Federated Vehicle Trajectory Prediction}\label{traj_pred}

\subsection{Dataset}\label{dataset}
We utilize the Zenseact Open Dataset (ZOD)~\cite{alibeigi2023zenseact}, a multi-modal autonomous driving dataset collected over a period of 2 years across 14 European countries.  
The dataset contains three subsets: \emph{frames} that are primarily suitable for non-temporal perception tasks, \emph{sequences} that are intended for spatio-temporal learning and prediction, and \emph{drives} that are aimed at longer-term tasks such as localization, mapping, and planning.
The \emph{frames} consists of more than 100k traffic scenes that have been carefully curated to cover a wide range of real-world driving scenarios.
From the original 100K images in the ZOD-dataset, only 80k images were usable after filtering for missing, incomplete, or erroneous data.
For each frame, the dataset contains annotations, calibration data, blurred and Deep Natural Anonymization Technology (dnat) images, ego-motion data, lidar data, and metadata on driving conditions. 
In the experiments, only blurred images were used.

Each image is associated with GNSS/IMU data that provides reliable navigation and positioning information. 
We shall focus on the task of vehicle trajectory planning and leverage the positioning information to automatically anotate the image frames as in~\cite{viala2023towards}. 
The ground truth is constructed by interpolating 17 points from the GNSS/IMU data, 3D-points in the trajectory from the original position of the car. 
The target distances of the 17 points are given by $\lbrace t_i \rbrace_{i=1}^{17}$ where $t_i = 5i$ for $1\leq i\leq 8$, $t_i=10(i-8)+40$ for $i \leq 12$, and $t_i=15i(i-12)+80$ $i > 12$.
Hence, the annotations emphasizes accuracy in the predicted trajectory close to the ego vehicle. 

\begin{figure}
\centering
\includegraphics[width=1\linewidth]{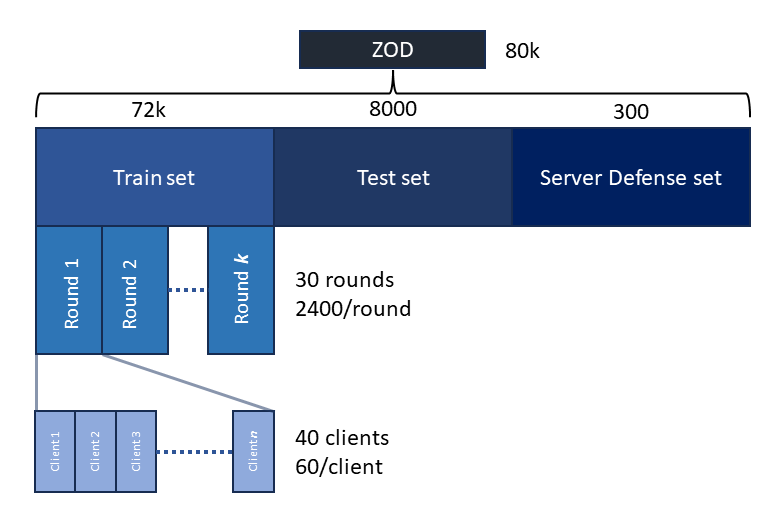}
\caption{Visual representation of the dataset split, illustrating the number images of the ZOD-dataset that were used for training (and how they are partitioned among clients), testing, and server defense.}
\label{fig:dataset_split}
\end{figure}

The dataset is split into a training, test and a server defense set, as seen in Fig.~\ref{fig:dataset_split}. 
To facilitate federated learning, the training set is further divided into separate sets for each global round and client. 
This partitioning is different from vanilla federated learning where the dataset remains static at each client.
In self driving, however, the car may be unable to store the data locally and must, hence, discard some of the data to make room for new. 
We capture this behavior by replacing the local data of all clients in every training round.
The test set is used to evaluate the model after each round. 
For the \textsc{OTA}, a test set was also created by including the backdoor trigger pattern in each image, leaving the ground-truth trajectory unchanged, to assess the attack success. 
Finally the server defense set may be used in conjunction with mitigation strategies employed by the server during training.

\subsection{Vehicle Trajectory Prediction} 
We employ the MobileNet-V3~\cite{howard2019searching} as the backbone of the trajectory prediction, pretrained on the ImageNet dataset~\cite{imagenet}.
MobileNet-V3 is a convolutional neural network optimized for mobile phone CPUs. 
We replace the head of network by 3 linear layers: 1024 neurons with ReLU activation, 512 neurons with ReLU activation, and 51 neurons without activation function. 
The 51 neurons in the final layer correspond to the 17 three-dimensional points $\lbrace \hat{p}_i \rbrace_{i=1}^{17}$, $\hat{p}_i\in\mathbb{R}^3$, representing the predicted trajectory. 
To facilitate the learning, we let $\hat{p}_{ij} \in [0,1]$, $j\in[3]$, and multiply $\hat{p}_{ij}$ with $t_i$, to obtain the point's position relative to the vehicle.
This allows the network to treat each predicted point equally. 

During training, we employ the Adam optimizer with a learning rate of 0.001, a batch size of 32, and the $\mathrm{L}1$-loss function.
Hence, for a given data point, consisting of an image $x$ and a ground-truth trajectory $\lbrace p_i \rbrace_{i=1}^{17}$, the loss is obtained as
\begin{equation}
    \ell(x, \lbrace p_i \rbrace_{i=1}^{17}; \theta)= \frac{1}{17} \sum_{i=1}^{17} \lVert p_i - \hat{p}_i \rVert_1  \label{eq1}
\end{equation}
where $\lbrace \hat{p}_i \rbrace_{i=1}^{17} = \theta(x)$ is the predicted trajectory.


\subsection{Federated Learning}
For the federated learning, we consider a network consisting of 40 clients.
The training is performed over 30 global training rounds where each round consists of 3 local epochs. 
As already mentioned, the clients are assumed to have collected a new dataset in the beginning of each training round.
This is illustrated in Fig.~\ref{fig:dataset_split} where the 72K training samples are split over the 30 training rounds and then, within each training round, further split over the 40 clients resulting in 60 data points per client.
Note that the data partitioning is performed randomly.
Although a random data partitioning is not realistic, e.g., consecutive data frames have a strong correlation in environment and weather, such partitioning was not feasible at the time of writing and left as an interesting future direction of study.

We assume that 4 out of the 40 clients are malicious.
Furthermore, during the federation, the server randomly samples 10 out of the 40 clients in each round.
Hence, the prevalence of malicious users may vary between 0\% to 40\% in a given training round.
The aggregation at the server is achieved by federated averaging~\cite{mcmahan2017communication}.
Pseudo code for the federated learning procedure is provided in Algorithm \ref{algo}.

\begin{algorithm}[t]
\caption{Federated Learning Procedure} \label{algo}
\begin{algorithmic}[1]
\State \textbf{Server side}
\State $\theta_{\mathrm{global}}$ $\gets$ pretrained MobileNetV3
\For{r from 1 to 30}
    \State {$S_{r}$ $\gets$ 10 clients selected at random}
    \State {Broadcast $\theta_{\mathrm{global}}$ to $S_r$}
    \For {client $c \in  S_{r}$}
        \State{ $\theta_c \gets $ TrainClient($\theta_{\mathrm{global}}$, $r$)}
    \EndFor
    \State $\theta_{\mathrm{global}}$ $\gets$ Aggregate($\theta_{\mathrm{global}}$, $\lbrace \theta_c \rbrace_{c\in S_r}$)
\EndFor
\State
\State \textbf{Client side}

\Function{TrainClient}{$\theta_{\mathrm{global}}$, $r$}
    \State $D_{r} \gets $ get dataset for current client and round
    \State {$\theta_{\mathrm{client}}$ $\gets$ $\theta_{\mathrm{global}}$}
    \For{each epoch $e$ from 1 to 3}
        \For{each batch $b \in D_r$}
            \State {Update $\theta_{\mathrm{client}}$ using $b$}
        \EndFor
    \EndFor
    \State \Return $\theta_{\mathrm{client}}$
\EndFunction
\end{algorithmic}
\end{algorithm}

\section{Experiments}\label{result}

\begin{table*}[t!]
\caption{Training score of the resulting model in conjunction with a given attack-defense combination.}
\label{table:10_avg_results}
    \centering
    \scalebox{0.9}{
    \begin{tabular}{ccccccccccc} 
        \toprule
         Attack Name & No-Defense & \textsc{Krum} & \textsc{Multi-Krum} & \textsc{LFR} & \textsc{FLTrust} & \textsc{PCA Defense} & \textsc{Loss Defense} & \textsc{Trimmed Mean} & \textsc{LossFusion} \\
        \midrule
        No Attack & 3.114&3.564&3.460&3.057&3.043&3.260&3.027&3.158&{2.990} \\
        
        Label-flipping & 7.924&3.381&3.446&3.058&3.616&3.397&3.071&4.015&{3.043} \\
       
        \textsc{Gradient Ascent} & 250.489&3.518&3.450&3.994&3.773&4.737&3.102&7.552&{3.030} \\
        
        \textsc{MSA} & 4.402&3.456&3.447&3.067&3.130&3.190&4.437&3.178&3.013 \\
        
        \textsc{FLStealth}  &$34.23\cdot 10^{10}$ &5.423&4.685&42.478&483.935&$21.91\cdot 10^{8}$&3.025&$32.63\cdot 10^{5}$&3.086 \\
     \bottomrule
    \end{tabular}
    }
\end{table*}

In this section, we assess the robustness of FL using various poisoning attacks and defense strategies. 
The experiments were performed on a single NVIDIA Quadro RTX5000 GPU with 8 cores, 40GB RAM and 500GB disk space.  
The duration of one experiment on the entire dataset is 20-30 minutes.

\subsection{Untargeted Attacks} \label{untargeted}
To measure the outcome from the federated training, the test loss of the global model is averaged over the last 10 training rounds, we refer to this metric as \textit{training score}. 
A high training score indicates a global model with poor performance, potentially due to a successful attack. 
On the other hand, a good model yields a low training score, possibly due to a weak attack or of a successful defense.
Moreover, we report the training scores as the average over 10 separate runs, i.e., each (attack, defense) combination is executed 10 times.

We consider 5 different poisoning attacks, including our novel \textsc{FLStealth} attack, and 8 different mitigation strategies.
As a baseline, we also provide the result without any mitigation strategies referred to as No-Defense.
For attacks requiring parameters, we consider: 1) in the label flipping ground truth trajectories are multiplied by -100, 2) for \textsc{MSA}, we shuffle 100 random rows in the weight matrix of each linear layer, 3) for \textsc{FLStealth}, the byzantine model is trained for 15 epochs using a learning rate of 0.0001 and $\kappa=10^{-9}$.
Note that a small value of $\kappa$ is typically required as the mean-squared error between the honest and byzantine models is in general much smaller than the loss.
Similarly, for defenses requiring parameters, we use: 1) in Krum, we use $4$ byzantine clients, 2) in Multi-Krum, we use $4$ byzantine clients and $6$ models to be aggregated, 3) in Trimmed Mean, after ordering the client updates based on magnitude, two clients are removed from the bottom and from the top of the ordering, and 4) for \textsc{PCA defense}, \textsc{LFR}, \textsc{Loss Defense}, and \textsc{LossFusion}, 4 clients are excluded in each round.
Note that the parameters are chosen in favor of the defenses as the correct number of malicious clients from the entire client set is used.

The \textsc{LossFusion} defense mehchanism is a simple fusion of \textsc{LFR} and \textsc{Loss Defense} after running them separately.
In particular, let $\theta_{\mathrm{LFR}}$ and $\theta_{\mathrm{LD}}$ denote the resulting model parameters after employing the two defense mechanisms separately.
Then, \textsc{LossFusion} selects the model parameters as
\[
\theta_{\mathrm{LF}} = 
\begin{cases}
\theta_{\mathrm{LFR}} & \text{for } \ell(D_\mathrm{server};\theta_{\mathrm{LFR}}) < \ell(D_\mathrm{server};\theta_{\mathrm{LD}})\\
\theta_{\mathrm{LD}} & \text{otherwise }
\end{cases}
\]
where $\ell(D_\mathrm{server}; \theta)$ is the average loss on the server's defense dataset using a model $\theta$.  
\textsc{LossFusion} aims at alleviating the weakness of only considering pre-aggregation losses in \textsc{Loss Defense} and of only looking at post-aggregated losses in \textsc{LFR}.
Hence, \textsc{LossFusion} effectively eliminates attacks targeting either \textsc{LFR} or \textsc{Loss Defense} since now both defenses must be bypassed.

In Table \ref{table:10_avg_results}, we illustrate the average training score for each attack-defense combination.
It can be seen that some combinations, particularly involving \textsc{FLStealth}, results in very high training scores.
The reason for this is that some of the attacks can be made arbitrary strong when able to bypass the defense.
Among the attacks, \textsc{FLStealth} achieves the largest training score for all defenses but the \textsc{LossDefense}.
On the other hand, among the defenses, \textsc{LossFusion} achieves the lowest training score on all attacks but \textsc{FLStealth}.

\subsection{Targeted Attacks} \label{results:backdoors}

The design of our targeted attack, \textsc{OTA}, involves three steps: 1) how to inject a trigger to an image, 2) how to alter the ground truth trajectory, and 3) decide how large portion of the data to poison.

\subsubsection{Trigger Injection}
Although there are many ways to design a trigger, in this paper, a simple square pattern was chosen.
Based on this choice, multiple features were studied, e.g., size, color, and total number of squares added.
Empirically, position and size surfaced as the main factors for a successful attack; varying the color of the square between red, green and white, or increasing the number of squares did not affect the overall performance of \textsc{OTA}.
Hence, for simplicity, only one red square were used for the final experiments.

To understand the impact of the square's position, experiments were conducted positioning it at the top-left corner, the center of the image, or at a random position for each image in the byzantine dataset. 
From these experiments, random position often went unnoticed by the defenses and hence that option was used for further experiments. 
However, we remark that positioning the square in the center performed the best but was deemed unrealistic, see Section~\ref{Backdoor_discussion}).

Finally, the size of the square only matters when it gets too small for the network to notice. 
The size was set as a percentage of the height of the image and performance dropped at around 5\% of the height. 
Sizes of up to 16\% of the image height was used with success, and for consistency in further experimentation a size of 10\% was used. 

\subsubsection{Altering the Ground-Truth Trajectory}
When a trigger is injected to a data sample, the corresponding ground-truth trajectory should also be modified in order to change the behavior of the model. 
We considered three such modifications: 1) make the car turn by the end of its path, 2) make the car go straight, and 3) make the car sig-sag around the ground-truth trajectory. 
From experimenting, the attack was deemed successful only when the car was made to turn, hence, for the final experiments, a trigger will force the car to turn.

It should be noted that a turn change can be achieved in several ways, e.g., by changing the angle of the turn, the sharpness of the turn, or the direction (left/right).
As most variations demonstrated similar result, a set-up with a turn to the right by modifying the last 5 points of the ground truth was chosen.

\subsubsection{Number of Poisoned Examples}
\begin{figure}
\centering
\includegraphics[width=0.5\textwidth]{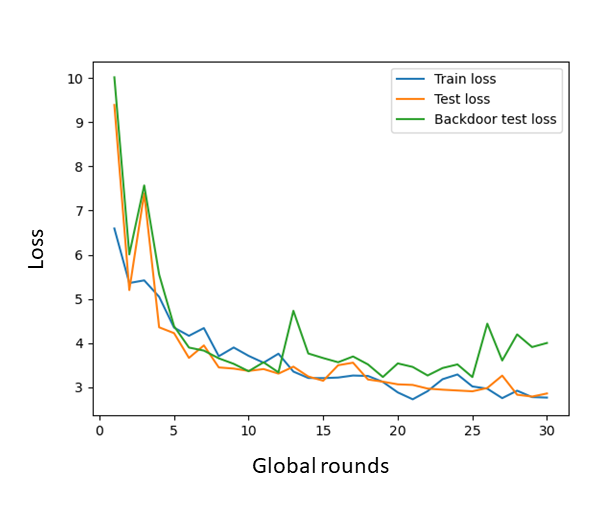}
\caption{Loss trajectories for a successful targeted attack (\textsc{OTA}) against \textsc{LFR} defense.}
\label{backdoor_loss}
\end{figure}

\begin{figure*}
\centering
\begin{subfigure}{0.3\textwidth}
    \includegraphics[width=\textwidth]{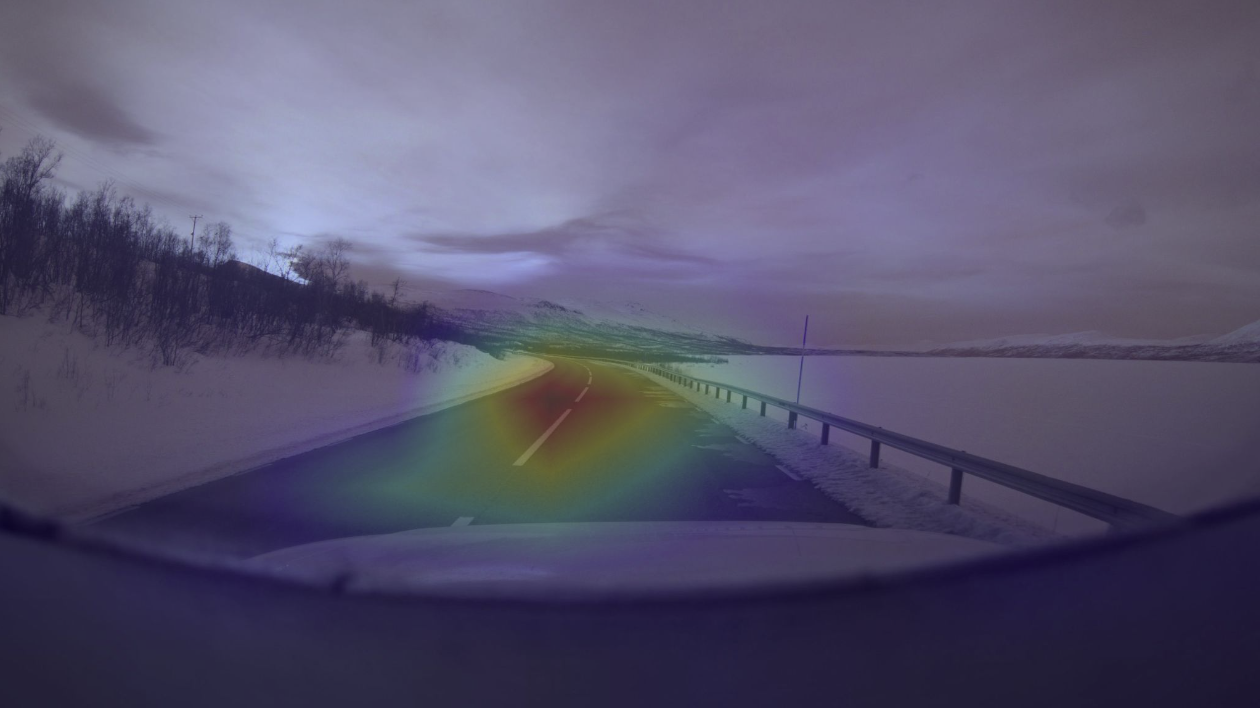}
    \caption{Attention heatmap of global model on a normal image.}
    \label{fig:heatmap_center}
\end{subfigure}
\hfill
\begin{subfigure}{0.3\textwidth}
     \includegraphics[width=\textwidth]{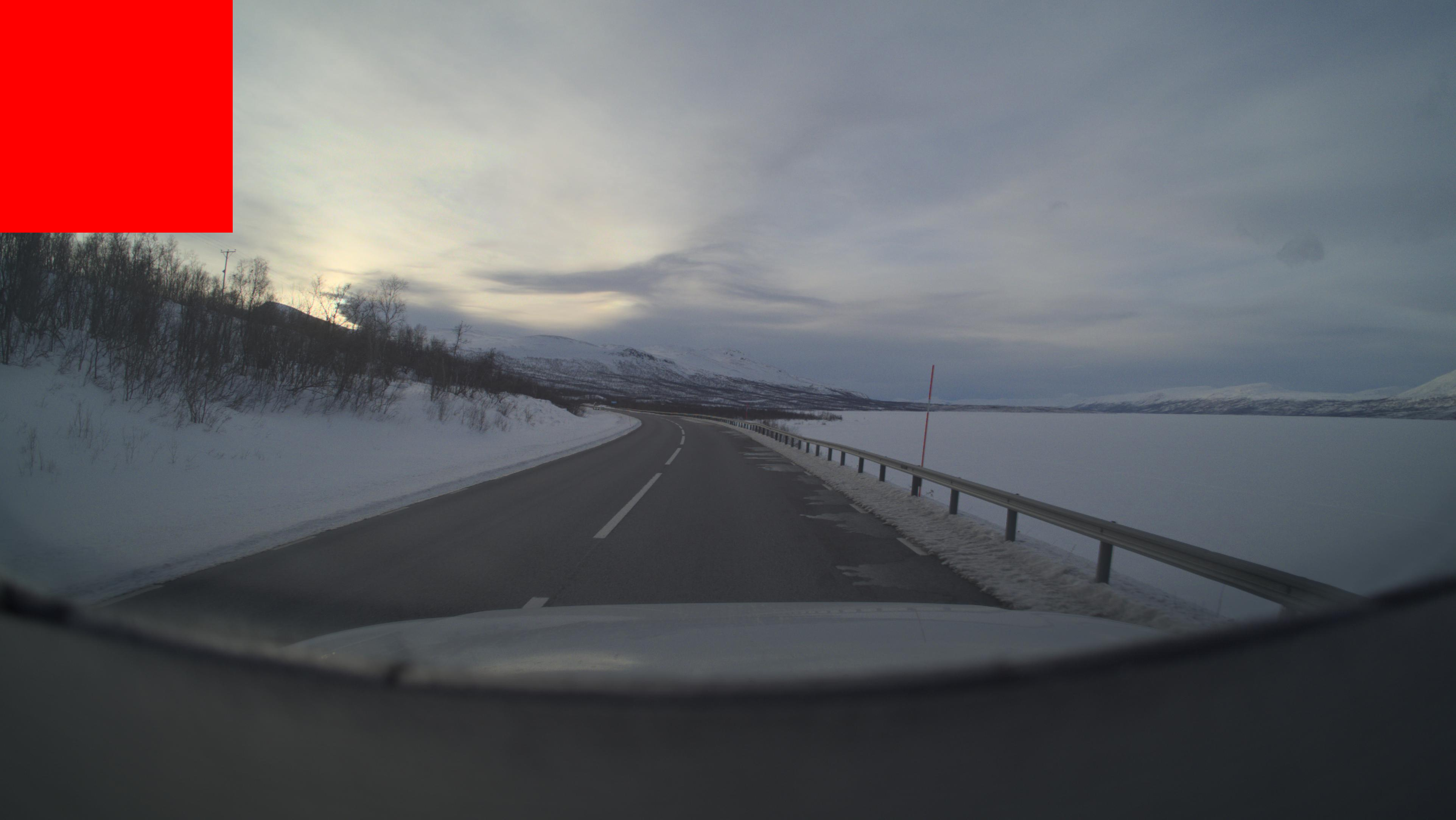}
     \caption{Trigger injected (a red square) in the top left corner.}
     \label{fig:square_in_tl_corner}
\end{subfigure}
\hfill
\begin{subfigure}{0.3\textwidth}
     \includegraphics[width=\textwidth]{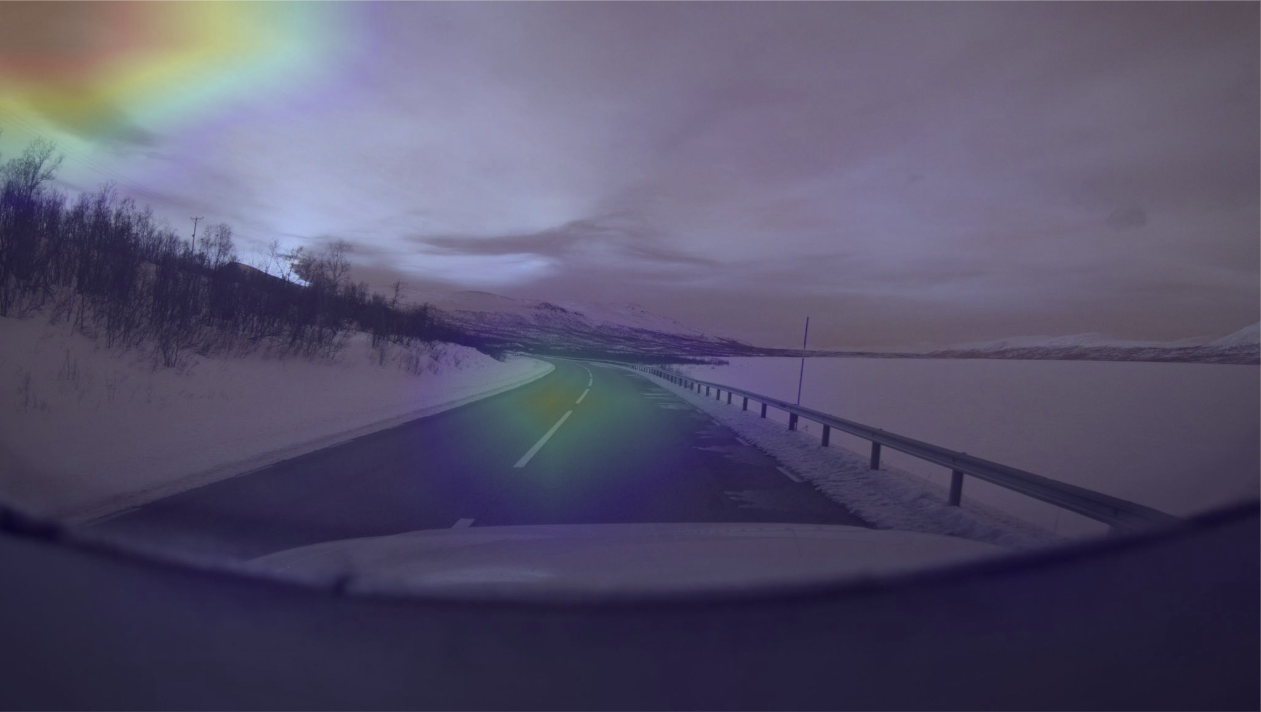}
     \caption{Attention heatmap of global model after backdoor is added to the image.}
     \label{fig:Heatmap_TL_corner}
\end{subfigure}
\caption{Change of model attention when a backdoor is added to the picture (Frame \#074220 in the ZOD-dataset)}
\label{fig:heatmaps}


\end{figure*}

The final component of the \textsc{OTA} is to choose the amount of data samples to poison.
From experiments with 20\% to 100\% of the data samples being poisoned, a trade-off was identified. 
A too large portion resulted in the backdoor becoming ineffective as the trigger is mostly present resulting in the entire dataset being poisoned and, consequently, the client model being easily identified as malicious. 
On the other hand, a small portion of poisoned data resulted in the model not learning the trigger at all.
Empirically, we found that a portion of 30\% of the dataset being poisoned yielded good results.
In Fig.~\ref{backdoor_loss} the loss trajectories are illustrated for a successful targeted attack.
From the test loss trajectory on the backdoor test set, see Section~\ref{dataset}, we notice that the loss trajectory increases by the end of the learning procedure which indicates a successful attack, i.e., the predicted trajectory deviates from the ground-truth trajectory in the presence of a trigger.
Another way of visualizing a successful backdoor attack is by the attention heat maps, as shown in Fig.~\ref{fig:heatmaps}. 
The series of images shows how the attention of the model is shifted from the road to the top left corner after the attack.

\subsubsection{Results}
\begin{table}[t]
\caption{Training and backdoor score from an \textsc{OTA}.}
\label{table:2}
    \centering
    \begin{tabular}{cccc}
        \toprule
        Defense &Training score & Backdoor score & Difference \\
        \midrule
        No defense & 3.19 & 3.52 & 0.33  \\
        \textsc{LFR}  & 2.92 & 3.31  & 0.33\\
        \textsc{Loss Defense} & 2.99 & 3.18 & 0.19\\
        \textsc{PCA Defense} & 3.27 & 3.48  & 0.21\\ 
        \textsc{Multi-Krum} & 3.24 & 3.30 & 0.06\\ 
        \textsc{FLTrust} & 3.01 & 3.40 & 0.39 \\
        \textsc{LossFusion} & 3.08 & 3.28 & 0.20 \\
     \bottomrule
    \end{tabular}
\end{table}

To measure the success of \textsc{OTA}, we consider both the training score, similar to untargeted attacks, but also a metric called backdoor score, computed similarly to the training score but over the backdoor test dataset, i.e., the same test set as in the training score but with triggers injected in images.
We expect a successful \textsc{OTA} to achieve a low training score, i.e., perform well on images without triggers, while simultaneously achieving a large backdoor score, i.e., deviate from ground-truth trajectories when triggers are present.
Table \ref{table:2} illustrates the performance of \textsc{OTA} against six defense mechanisms with parameters chosen as in Section~\ref{untargeted}.  
The difference between the training score and the backdoor score indicates the effectiveness of the attack with a larger difference yielding a more successful attack.
The values in each row in Table~\ref{table:2} is the average over 5 independent runs. 

From Table~\ref{table:2}, it can be seen that \textsc{Loss Defense} and \textsc{LossFusion} are effective at mitigating \textsc{OTA}, yielding a difference of 0.19 and 0.20, respectively. 
Although \textsc{Multi-Krum} displays the lowest difference of 0.06, the training score is large. 
Visual inspection of predictions obtained from models trained with \textsc{Multi-Krum} mitigation also entails that the model is poisoned, i.e., predictions follow the expected behavior when exposed to the trigger. 

\begin{figure*}
\centering
\begin{subfigure}{0.45\textwidth}
    \includegraphics[width=\textwidth]{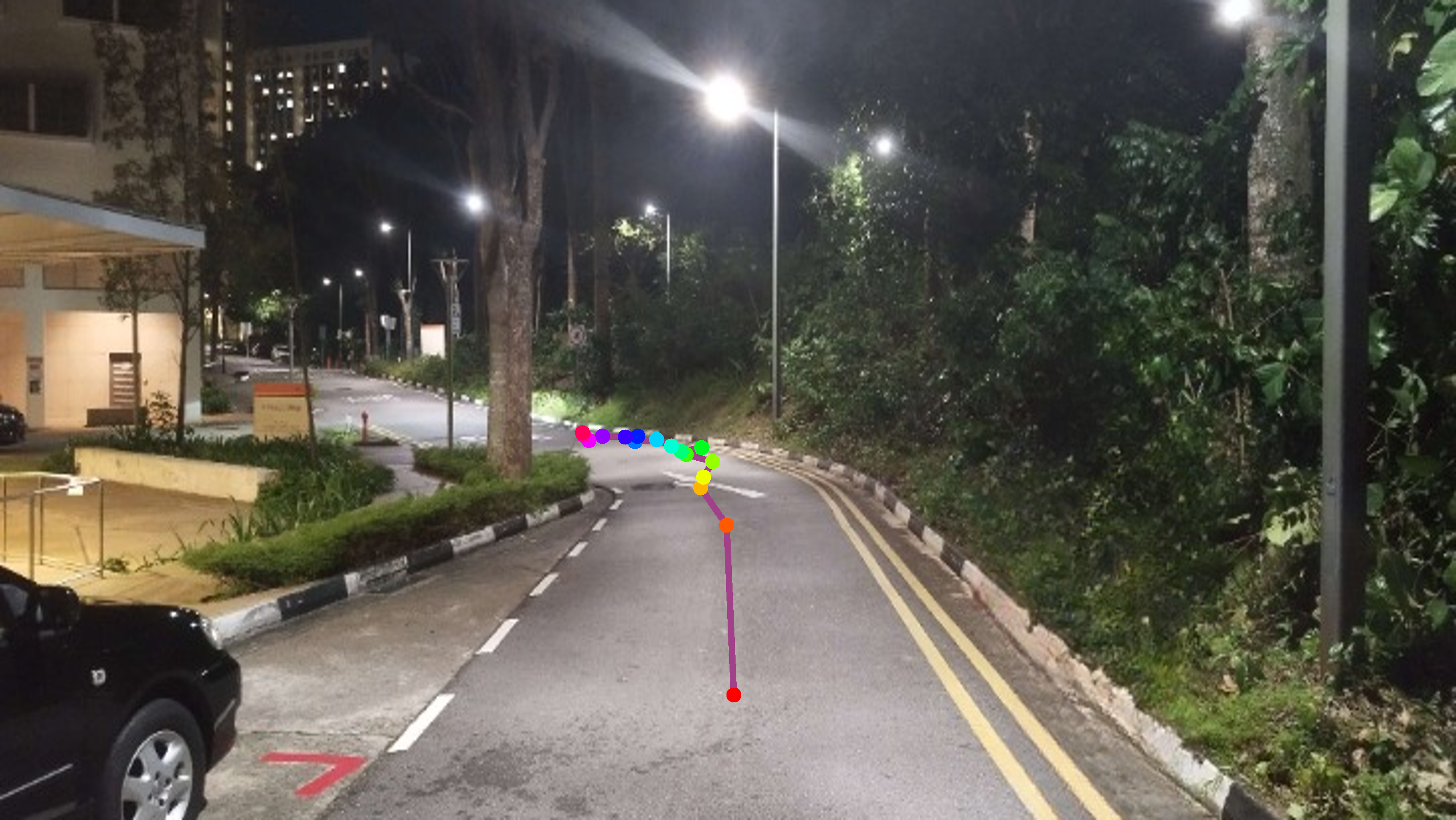}
    \caption{Trajectory prediction on road in Singapore\\~}
    \label{fig:pred_in_madison}
\end{subfigure}
\hfill
\begin{subfigure}{0.45\textwidth}
     \includegraphics[width=\textwidth]{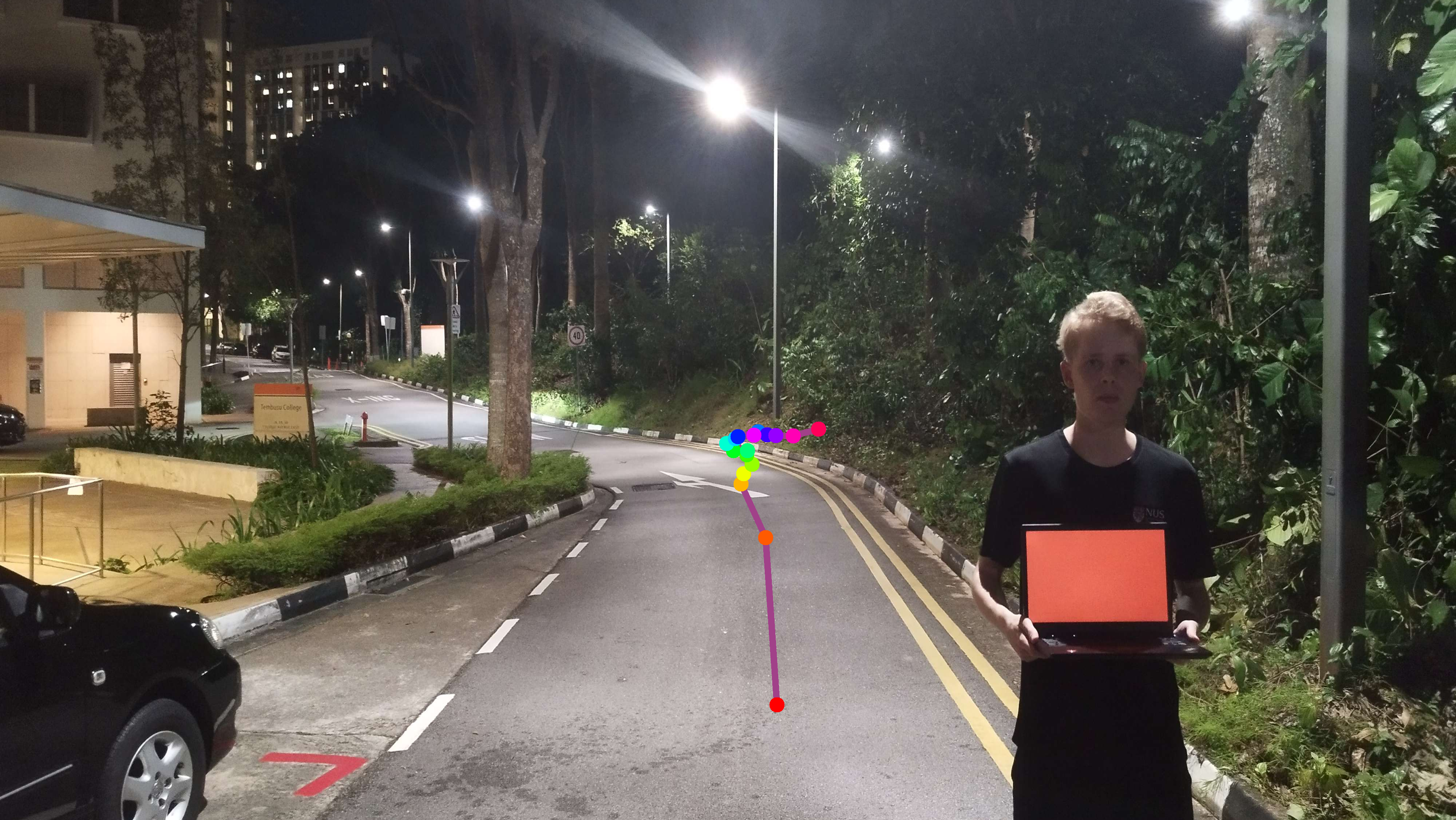}
     \caption{The same image as Fig.~\ref{fig:pred_in_madison} but with a malicious actor showing the trigger pattern.}
     \label{fig:backdoor_in_madison}
\end{subfigure}
\caption{OTA performed in practice.}
\label{fig:real_backdoor}



\end{figure*}

To further test the robustness of \text{OTA}, an attack was performed in a more realistic setting, as shown in Fig.~\ref{fig:real_backdoor}. 
Fig.~\ref{fig:backdoor_in_madison} displays a road with a person showing a trigger pattern on a computer screen. 
A photo without the person was then generated, using image processing tools, in order to keep the environment fixed. 
The model, subject to the \textsc{OTA}, employing the \textsc{Loss Fusion} defense was then used to predict the trajectory on each image. 
Without the trigger pattern the model produces a reasonable prediction of the trajectory, see Fig.~\ref{fig:pred_in_madison}, and when the backdoor pattern was introduced, the model sends the car to the right, see Fig.~\ref{fig:backdoor_in_madison}, which, in this case, is the opposite of the intended direction.

\section{Discussion}
In the following section, we discuss our results pertaining to untargeted and targeted attacks, respectively.

\subsection{Untargeted Attacks} \label{FLStealth_discussion}
The \textsc{FLStealth} attack aims to deteriorate the global model while remaining undetected. 
This proves to be effective against all considered defenses, apart from \textsc{Loss Defense} and \textsc{LossFusion}. 
For \textsc{Krum}, \textsc{Multi-Krum}, \textsc{FLTrust} and \textsc{PCA Defense}, these results are expected as they rely on a similarity score for each client and mitigate the impact of dissimilar clients. 
Since \textsc{FLStealth} is designed to provide poisoned models similar to those of honest clients, the malicious clients will have a similar similarity score to an honest client.
\textsc{FLStealth} is also expected not to bypass \textsc{Loss Defense} as it is designed to increase the loss which is exactly the signal that \textsc{Loss Defense} operates on. 
Decreasing $\kappa$ will improve the chances of bypassing also \textsc{Loss Defense} but will also reduce the effect of the attack. 
 
Interestingly, \textsc{FLStealth} and the related \textsc{Gradient Ascent} attack both perform well against \textsc{LFR}. 
We observe that this is because the attack sometimes, but rarely, bypasses \textsc{LFR} completely. 
For each such instance, at least two attackers are present and removing one of them results in a worsened model. 
This counter intuitive phenomenon is due to the inner workings of \textsc{LFR} that removes clients sequentially based on the loss impact of each client.
When multiple attackers are present, their updates may partially cancel out and may, in some cases, result in a low loss when both are included but an increased loss when one is removed. Since \textsc{LFR} does not take into account the relationship between multiple clients, the defense will not realise that the best strategy is to remove both clients but will, instead,  remove 4 other clients, amplifying the attack further since it now contributes more to the averaged model.

\subsection{Targeted Attacks} \label{Backdoor_discussion}
OTA successfully evades all the defenses, hence poisoning the global model and injecting the trigger into all vehicles in the federation. 
Since the model is trained to make good predictions when no trigger pattern is present the targeted model will have low loss. 
This is the reason why loss-based mitigation strategies are unsuccessful. 
The second category of defenses focus on the similarity of the received client gradients. However, as the malicious clients only poison 30\% of their local data, their updates will be similar to that of a benign client, rendering similarity-based defenses ineffective. 

During the experiments, some defenses were sometimes able to counter or cancel out \textsc{OTA} in a single training round. 
However, if the malicious client manages to bypass the defense in only a single round, the trigger will be present for all clients going forward. 
This may further allow the attacker to bypass the defense in future rounds, amplifying the effect of the attack.

As mentioned in Section~\ref{results:backdoors}, there are several ways of adding trigger patterns. 
The empirical results suggested that the best positioning for a trigger pattern is in the center of the image. 
This is expected since that square would cover the most important part of the image, where the model's attention is focused, i.e., the road. 
However, in real life this would limit the position of the attacker and make the attack more difficult to execute, hence, this positioning was rejected.

\section{Conclusion}\label{concl}
This paper studies vulnerabilities of federated learning applied in the area of regression tasks within autonomous driving. 
We have introduced two novel attacks: 1) an untargeted attack called \textsc{FLStealth} tailored to deteriorate the global model while remaining stealthy and 2) a targeted attack \textsc{OTA} aiming to inject triggers to make the car turn when exposed to the trigger.
A thorough assessment of the attack success was performed by comparing to other types of attacks and to common poisoning mitigation strategies in federated learning.

Our results have highlighted the significant threat posed by backdoor attacks, calling for effective detection methods and exploring ensemble techniques that combine different approaches that could enhance defenses against targeted attacks. 
Notably, we observed that none of the existing defenses effectively countered \textsc{OTA}. 
Finally, we demonstrated the benign effects of combining multiple defensive strategies, as demonstrated by the introduced \textsc{LossFusion} defense.

\bibliographystyle{ieeetr}
\bibliography{main} 

\begin{thebibliography}{10}

\bibitem{bojarski2016end}
M.~Bojarski, D.~Del~Testa, D.~Dworakowski, B.~Firner, B.~Flepp, P.~Goyal, L.~D. Jackel, M.~Monfort, U.~Muller, J.~Zhang, {\em et~al.}, ``End to end learning for self-driving cars,'' {\em arXiv:1604.07316}, 2016.

\bibitem{GDPR}
{European Union}, ``General data protection regulation {(GDPR)} information portal,'' 2023.
\newblock Available at: \url{https://gdpr-info.eu}.

\bibitem{mcmahan2017communication}
B.~McMahan, E.~Moore, D.~Ramage, S.~Hampson, and B.~A. y~Arcas, ``Communication-efficient learning of deep networks from decentralized data,'' in {\em AISTATS}, 2017.

\bibitem{aparna2021steering}
M.~Aparna, R.~Gandhiraj, and M.~Panda, ``Steering angle prediction for autonomous driving using federated learning: the impact of vehicle-to-everything communication,'' in {\em IEEE Int. Conf. on Comp. Comm. and Netw. Technologies (ICCCNT)}, 2021.

\bibitem{savazzi2021opportunities}
S.~Savazzi, M.~Nicoli, M.~Bennis, S.~Kianoush, and L.~Barbieri, ``Opportunities of federated learning in connected, cooperative, and automated industrial systems,'' {\em IEEE Communications Magazine}, vol.~59, no.~2, 2021.

\bibitem{nguyen2022deep}
A.~Nguyen, T.~Do, M.~Tran, B.~X. Nguyen, C.~Duong, T.~Phan, E.~Tjiputra, and Q.~D. Tran, ``Deep federated learning for autonomous driving,'' in {\em IEEE Intelligent Vehicles Symposium (IV)}, 2022.

\bibitem{FedLearnIR}
Y.~Chen, C.~Wang, and B.~Kim, ``Federated learning with infrastructure resource limitations in vehicular object detection,'' in {\em IEEE/ACM Symposium on Edge Computing (SEC)}, 2021.

\bibitem{TurnSigFed}
D.~S., K.~N., and S.~Athavale, ``Turn signal prediction: A federated learning case study,'' {\em arXiv 2012.12401}, 2020.

\bibitem{OrchFed}
D.~Deveaux, T.~Higuchi, S.~Uçar, C.-H. Wang, J.~Härri, and O.~Altintas, ``On the orchestration of federated learning through vehicular knowledge networking,'' in {\em IEEE Vehicular Networking Conference (VNC)}, 2020.

\bibitem{blanchard2017machine}
P.~Blanchard, E.~M. El~Mhamdi, R.~Guerraoui, and J.~Stainer, ``Machine learning with adversaries: Byzantine tolerant gradient descent,'' {\em Neurips}, vol.~30, 2017.

\bibitem{cao2020fltrust}
X.~Cao, M.~Fang, J.~Liu, and N.~Z. Gong, ``{FLtrust}: Byzantine-robust federated learning via trust bootstrapping,'' {\em arXiv 2012.13995}, 2020.

\bibitem{huang2020dynamic}
A.~Huang, ``Dynamic backdoor attacks against federated learning,'' {\em arXiv 2011.07429}, 2020.

\bibitem{sun2021data}
G.~Sun, Y.~Cong, J.~Dong, Q.~Wang, L.~Lyu, and J.~Liu, ``Data poisoning attacks on federated machine learning,'' {\em IEEE Internet of Things Journal}, vol.~9, no.~13, 2021.

\bibitem{li2021backdoor}
X.~Li, G.~Kesidis, D.~J. Miller, and V.~Lucic, ``Backdoor attack and defense for deep regression,'' {\em arXiv:2109.02381}, 2021.

\bibitem{wang2023bandit}
S.~Wang, Q.~Li, Z.~Cui, J.~Hou, and C.~Huang, ``Bandit-based data poisoning attack against federated learning for autonomous driving models,'' {\em Expert Systems with Applications}, vol.~227, 2023.

\bibitem{alibeigi2023zenseact}
M.~Alibeigi, W.~Ljungbergh, A.~Tonderski, G.~Hess, A.~Lilja, C.~Lindstr{\"o}m, D.~Motorniuk, J.~Fu, J.~Widahl, and C.~Petersson, ``Zenseact open dataset: A large-scale and diverse multimodal dataset for autonomous driving,'' in {\em IEEE/CVF International Conference on Computer Vision}, 2023.

\bibitem{mahloujifar2019universal}
S.~Mahloujifar, M.~Mahmoody, and A.~Mohammed, ``Universal multi-party poisoning attacks,'' in {\em ICML}, 2019.

\bibitem{guerraoui2018hidden}
R.~Guerraoui, S.~Rouault, {\em et~al.}, ``The hidden vulnerability of distributed learning in byzantium,'' in {\em ICML}, 2018.

\bibitem{xie2019dba}
C.~Xie, K.~Huang, P.-Y. Chen, and B.~Li, ``Dba: Distributed backdoor attacks against federated learning,'' in {\em ICLR}, 2019.

\bibitem{bagdasaryan2020backdoor}
E.~Bagdasaryan, A.~Veit, Y.~Hua, D.~Estrin, and V.~Shmatikov, ``How to backdoor federated learning,'' in {\em AISTATS}, 2020.

\bibitem{munoz2017towards}
L.~Mu{\~n}oz-Gonz{\'a}lez, B.~Biggio, A.~Demontis, A.~Paudice, V.~Wongrassamee, E.~C. Lupu, and F.~Roli, ``Towards poisoning of deep learning algorithms with back-gradient optimization,'' in {\em AISec}, 2017.

\bibitem{tolpegin2020data}
V.~Tolpegin, S.~Truex, M.~E. Gursoy, and L.~Liu, ``Data poisoning attacks against federated learning systems,'' in {\em ESORICS}, 2020.

\bibitem{shejwalkar2022back}
V.~Shejwalkar, A.~Houmansadr, P.~Kairouz, and D.~Ramage, ``Back to the drawing board: A critical evaluation of poisoning attacks on production federated learning,'' in {\em IEEE Symposium on Security and Privacy (SP)}, 2022.

\bibitem{baruch2019little}
G.~Baruch, M.~Baruch, and Y.~Goldberg, ``A little is enough: Circumventing defenses for distributed learning,'' {\em Neurips}, vol.~32, 2019.

\bibitem{fang2020local}
M.~Fang, X.~Cao, J.~Jia, and N.~Gong, ``Local model poisoning attacks to $\{$Byzantine-Robust$\}$ federated learning,'' in {\em USENIX}, 2020.

\bibitem{shejwalkar2021manipulating}
V.~Shejwalkar and A.~Houmansadr, ``Manipulating the byzantine: Optimizing model poisoning attacks and defenses for federated learning,'' in {\em Network and Distributed Systems Security Symposium}, 2021.

\bibitem{biggio2012poisoning}
B.~Biggio, B.~Nelson, and P.~Laskov, ``Poisoning attacks against support vector machines,'' {\em arXiv:1206.6389}, 2012.

\bibitem{yang2023model}
M.~Yang, H.~Cheng, F.~Chen, X.~Liu, M.~Wang, and X.~Li, ``Model poisoning attack in differential privacy-based federated learning,'' {\em Information Sciences}, vol.~630, 2023.

\bibitem{gu2019badnets}
T.~Gu, K.~Liu, B.~Dolan-Gavitt, and S.~Garg, ``Badnets: Evaluating backdooring attacks on deep neural networks,'' {\em IEEE Access}, vol.~7, 2019.

\bibitem{yin2018byzantine}
D.~Yin, Y.~Chen, R.~Kannan, and P.~Bartlett, ``Byzantine-robust distributed learning: Towards optimal statistical rates,'' in {\em ICML}, 2018.

\bibitem{valadi2023fedval}
V.~Valadi, X.~Qiu, P.~P.~B. de~Gusm{\~a}o, N.~D. Lane, and M.~Alibeigi, ``Fedval: Different good or different bad in federated learning,'' {\em USENIX}, 2023.

\bibitem{xhemrishi2023fedgt}
M.~Xhemrishi, J.~Östman, A.~Wachter-Zeh, and A.~G. i~Amat, ``{FedGT}: Identification of malicious clients in federated learning with secure aggregation,'' {\em arXiv:2305.05506}, 2023.

\bibitem{viala2023towards}
A.~Viala~Bellander and Y.~Ghafir, ``Towards federated fleet learning leveraging unannotated data,'' Master's thesis, Chalmers University of Technology, 2023.

\bibitem{howard2019searching}
A.~Howard, M.~Sandler, G.~Chu, L.-C. Chen, B.~Chen, M.~Tan, W.~Wang, Y.~Zhu, R.~Pang, V.~Vasudevan, {\em et~al.}, ``Searching for mobilenetv3,'' in {\em IEEE/CVF International Conference on Computer Vision}, 2019.

\bibitem{imagenet}
J.~Deng, W.~Dong, R.~Socher, L.-J. Li, K.~Li, and L.~Fei-Fei, ``Imagenet: A large-scale hierarchical image database,'' in {\em IEEE Conference on Computer Vision and Pattern Recognition}, 2009.

\end{thebibliography}

\end{document}